\documentclass[preprint,12pt]{elsarticle}

\usepackage{amssymb}

\journal{BICA Journal (to appear). Pre-print version}

\begin{document}

\begin{frontmatter}



\title{Artwork creation by a cognitive architecture integrating computational creativity and dual process approaches}


\author[ICARaddress]{Agnese Augello}
\author[ICARaddress]{Ignazio Infantino}
\author[UniTOaddress,ICARaddress]{Antonio Lieto}
\author[ICARaddress]{Giovanni Pilato}
\author[ICARaddress]{Riccardo Rizzo}
\author[ICARaddress]{Filippo Vella}

\address[ICARaddress]{ICAR-CNR, edificio 11, Viale delle Scienze, Palermo, Italy}
\address[UniTOaddress]{University of Turin, Dip. di Informatica,Corso Svizzera 185, Torino, Italy}

\address{To appear in the journal Biologically Inspired Cognitive Architectures (2016). Pre-print version.}

\begin{abstract}
The paper proposes a novel cognitive architecture (CA) for computational creativity based on the Psi model and on the mechanisms inspired by dual process theories of reasoning and rationality.  In recent years, many cognitive models have focused on dual process theories to better describe and implement complex cognitive skills in artificial agents, but creativity has been approached only at a descriptive level. In previous works we have described various modules of the cognitive architecture that allows a robot to execute creative paintings.  By means of dual process theories we refine some relevant mechanisms to obtain artworks, and in particular we explain details about the resolution level of the CA dealing with different strategies of access to the Long Term Memory (LTM) and managing the interaction between S1 and S2 processes of the dual process theory. The creative process involves both divergent and convergent processes in either implicit or explicit manner. This leads to four activities (exploratory, reflective, tacit, and analytic) that, triggered by urges and motivations, generate creative acts. These creative acts exploit both the LTM and the WM in order to make novel substitutions to a perceived image by properly mixing parts of pictures coming from different domains. The paper highlights the role of the interaction between S1 and S2 processes, modulated by the resolution level, which focuses the attention of the creative agent by broadening or narrowing the exploration of novel solutions, or even drawing the solution from a set of already made associations.  An example of artificial painter is described in some experimentations by using a robotic platform.

\end{abstract}

\begin{keyword}
Computational Creativity \sep Cognitive Architecture \sep Dual Process Theory \sep PSI model.



\end{keyword}

\end{frontmatter}


\section{Introduction}
\label{sec:Intro}

Cognitive architectures (CA) are inspired by functional mechanisms of human brain, and the various models proposed  \citep{goertzel2010world} try to define the necessary modules to emulate the complex interactions among perception, memory, learning, planning, and action execution. These modules influence the external behaviours of agents and their interactions with humans.
 
Among the many high-level cognitive abilities, creativity could be the one that allows to investigate new solutions or improvements for cognitive architectures for various reasons. 
Creativity, in fact, is a component of human intelligence that cannot ignore external influence, given that its mechanisms of association, analogy, and composition strongly depend  on some evaluation processes \citep{boden1998creativity}.

To provide a definition of 'what is creativity' is as difficult as for the definition of 'intelligence', and it is easier to focus on the final product of the creative process. Various attempts were made to formalize the artificial creativity (named also computational creativity) \citep{colton2011computational} \citep{Goel:2011:2069618} and to define methodologies and establish metrics of creativity \citep{galanter2012computational}. In particular, the research area on computational creativity focuses on computational systems capable to create artifacts and ideas, and it heavily relies on both internal and external system evaluations. Computational creativity research is then focused on the creation of an outcome, or a process, with a certain degree of innovation and invention, and capable of arousing emotions.


Many proposals have been presented in order to tackle the problem of computational creativity. For example, a cognitive model for creativity has been proposed in \citep{Gabora2002} with the aim of explaining the cognitive transformations that occur as the creative process is triggered and proceeds. In this model, after an initial, intuitive, phase which triggers an information retrieval process from memory, there is an analytical phase characterizing a focused form of thought which analyzes relationships of cause and effect. Two key points of this model are the variable focus of attention and the associative memory. In particular, the variable focus of attention, while pointing the basic idea, also collects other concepts that are parts of the stream of thought.  
Furthermore, by means of associations between different concepts and completion mechanisms, new and surprising results can emerge \citep{bogart2013}. This kind of creative process can be bound to the process that Boden calls \textit{combinatorial creativity} which is related to making unusual combinations, either consciously or unconsciously generated, of familiar ideas \citep{Boden2009}.

Cognitive architectures allow software to deal with problems that require contributions from both cognitive science and robotics, in order to achieve social behaviours which are typical of the human being. Currently, CA have had little impact on real-world applications and a limited influence in robotics. Their aim and long-term goal is the detailed definition of the Artificial General Intelligence (AGI) \citep{Goertzel_Pennachim_2007}, i.e. the construction of artificial systems that have a skill level equal to humans in generic scenarios.
Despite their current underutilization they represent, in our opinion, a possible general framework for the integration and analysis of creative computational processes in cognitive agents.  As a matter of fact, CA represent the infrastructure of an intelligent system that manages, through appropriate knowledge representation, the process of perception, recognition, categorization, reasoning, planning and decision-making \citep{Langley_etAl_2009}. It is a great open challenge to be able to integrate into the CA all the aspects relating to language, emotions, abstract thought, creativity and so on; and the main difficulties are due to the management of the internal representation of the external world, and the ability to efficiently update these models through the interaction with the outside world. 

In a general perspective, the \emph{mental} capabilities \citep{vernon2007} of artificial computational agents can be directly introduced into a cognitive architecture or emerge from the interaction of its components. In this respect, many approaches have been presented in the literature, ranging from cognitive testing of theoretical models of the human mind, to robotic architectures based on perceptual-motor components and purely reactive behaviours \citep{goertzel2010world}\citep{Langley_etAl_2009} \citep{cai2011openpsi}.

Various available cognitive architectures present some mechanisms of self-evaluation that can enable the comparison of expected results with the one obtained. However, as mentioned, in order to introduce creativity modules which can truly affect the creative process \citep{Besold2015} of a cognitive agent, it is necessary to capture external evaluation (e.g. feedbacks from users). 

Jordanous \citep{Jordanous} indicates 14 key components to take into account in order to deal with computational creativity, and some of them could be partially considered in various components of a CA: \emph{dealing with uncertainty, intention and emotional involvement, generation of results, domain competence, originality, social interaction and communication, variety, divergence and experimentation, value, evaluation}. Other factors are more difficult to manage, requiring more investigation and better definition for real implementation:         
 \emph{active involvement and persistence, general intellectual ability, progression and development, spontaneity and subconscious processing, independence and freedom, (abstract) thinking}. 
 Being capable to tackle at least some of these parameters can lead to realise an embodied artificial agent able to interact with humans and the environment, capable to produce a creative processes or creative acts trough a given cognitive architecture.








Along with these considerations, in previous works \citep{augello2013} we  presented the possibility of embedding some mechanisms of computational creativity \citep{Boden2009} \citep{Colton_etAl_2009} within cognitive architectures  \citep{goertzel2010world}\citep{Langley_etAl_2009}. 
We identified some features that have been exploited by a robot, capable of developing his own artificial visual perception, and originating a creative act based both on its experience and expertise, and through interaction with the human. 

As a consequence, we explored the features of the Psi model \citep{bartl1998psi}\citep{Bach:etAl:2006} and its architecture, since it explicitly involves  the concepts of emotion and motivation in cognitive processes, which are two important factors in creativity processes. In particular, we focused our approach on the MicroPsi \citep{Bach:etAl:2006}  architecture, which is an integrative architecture based on the Psi model that has been tested on some practical control applications. It has also been used on simulated artificial agents in a simple virtual world. 
The MicroPsi agent architecture is interesting since it describes the interaction of emotion, motivation and cognition of situated agents based on the Psi theory of Dietrich D\"orner. \citep{bach2009principles}\citep{Bach:etAl:2006}. MicroPsi consists of several main components, which are connected to their environment by a set of somatic parameters (named \emph{urges}) and external sensors. The activity of the agent consists of a number of internal and external behavior modules. While the former are ``mental'' actions, the latter send sequences of actuator commands to the environment. 

Recently, the dual process theories of mind (\citep{stanovich2000individual}, \citep{evans2009two}, \citep{kahneman2011thinking}) have suggested that our cognition is governed by two types of interacting cognitive systems, which are called respectively system(s) 1 and system(s) 2. Systems of the type 1, referred also as S1, operate with rapid, automatic, associative processes of reasoning. They are phylogenetically older and execute processes in a parallel and fast way. Type 2 systems, referred also as S2, are, on the other hand, phylogenetically more recent and are based on conscious, controlled, sequential processes (also called type 2 processes) and on logic-based rule following. As a consequence, if compared to system 1, system 2 processes are slower and cognitively more demanding.  In the dual process perspective, then, decision making consists in a two-step procedure based on the interaction between heuristic, perception-guided (and biased) thinking (type 1 processes), with forms of deliberative thinking based on the canons of normative rationality (and on type 2 processes).

In the area of computational creativity, despite initial attempts have been made to point out, at the descriptive level, the importance of the compatibility between such perspective and its implementation in a cognitive architecture involved in computational creativity task \citep{sun2015accounting}, at the best of our knowledge, there is not yet an existing implementation of an artificial system integrating a dual process account of computational creativity. 



In this paper we illustrate the evolution of the approach aimed at supporting the execution of an artificial digital painter \citep{augello2013}\citep{iccc2014}, which is based on the MicroPsi model. This evolution exploits the dual process theory to emulate creative processes. As in \citep{augello2013binding}\citep{iccc2014}, the proposed approach is based on a multilayer mechanism that implements an associative memory based on Self Organizing Maps (SOMs) \citep{Kohonen2001} and it is capable to properly mix elements belonging to different domains. 

The paper is structured as follows:  the next section discusses the dual process theory  and its application in the field of computational creativity, the third section proposes  a computational creative system  based on the PSI cognitive architecture and implementing the Dual Process theory, the fourth section illustrates some experimental results and final artworks, then in the last section  conclusions are given.

\section{Dual Process Approach  and Computational Creativity	}
\label{sec:dp}

In recent years, the cognitive modelling community have posed a growing attention on the dual process theories as a framework for modelling artificial cognition ``beyond the rational'' in the sense intended by Kennedy et al. \citep{kennedy_beyond_2012}. 

In this line, different attempts have been made to introduce the tenets of the dual process models in the computational arena. Efforts have been made, for example, in the areas of knowledge representation and reasoning \citep{frixione2014towards} \citep{lieto15ijcai}, cognitive systems dealing with arithmetical calculations \citep{strannegaard2013cognitive}, action selection \citep{Faghihi201538}, cognitive models of emotions \citep{larue12cognitive}, question answering \citep{lieto14knowledge} and in the design of general purpose cognitive architectures, such as CLARION, whose principles are explicitly inspired by such a theoretical framework \citep{sun2014interpreting}.  

A first theoretical attempt to apply the dual process theory in the field of computational modeling has been developed by Sloman \citep{sloman1996empirical} by adopting the Smolensky connectionist framework  \citep{smolensky1988proper} to describe the computational differences between system 1 and system 2. Interestingly, Sloman described the relationship between the two systems as parallel-competitive in nature, differently from the traditional default-interventionist approach, according to which the deliberative S2 reasoning processes work as logical controller and can inhibit the potentially biased responses of the S1 systems and replace them with correct outputs based on reflective reasoning (see (\citep{evans2009two}). More recently, also a revised version of the dual process theory, based on the hypothesis that the system 2 is divided in two further levels, respectively called ``algorithmic'' and ``reflective'' have also found a computational counterpart in the area of biologically inspired simulation of emotions emergence \citep{larue12cognitive}. 

Interestingly, the dual process approach has been considered plausibly applicable also in the area of creativity research \citep{sowden2014}. Creative thinking literature, in fact, distinguishes between two sets of processes: those involved in the generation of ideas, and those involved with their refinement, evaluation and/or selection. In addition, in this area, theoretical and experimental frameworks have been developed suggesting that, in human beings, the realisation of creative artefacts (either tangible or intangible) involves the interaction between the two types of reasoning processes hypothesised by the dual process approach \citep{sowden2014}. Such a perspective, then, seems to be also plausibly applicable to the design of novel models of computational creativity.

Nevertheless, reducing thinking processes as a straightforward dichotomy between two systems is judged to be too simplistic by many authors (see for example \citep{stanovich2000individual} and  \citep{Glockner} ). As a matter of fact, the key question is how the two systems interact. According to Sowden et al \citep{sowden2014}, the manner in which S1 and S2 cooperate influences the differences in creative capabilities arising both between individuals and across time with respect to the same individual.

Guilford, \citep{Guilford}  discriminated thinking processes as being divergent or convergent. The operation of these two processes can be seen as a precursor of the dual process model of creative thinking. Divergent processes are usually referred as being associative, while convergent processes are referred as being more analytic.\\
Nonetheless, in the literature it has been claimed that a straightforward mapping of divergent thinking onto S1 and convergent thinking onto S2 is too na\"ive, since divergent processes can be also the result of an associative course, but they can also be the result of a deliberative process, as well as convergent processes can be also unconscious.

This has led to the definition of disparate models of creativity, many of which related to different dual-process models. For example, Basadur et al. \citep{Basadur} distinguish between three major stages in the creative thinking process: \emph{a)} problem finding, \emph{b)} problem solving, and \emph{c)} solution implementation, by realizing a process that involves ideation-evaluation cycles. In this view, the creative thinking process is seen as a problem solving task thus involving different phases (this idea goes back to \citep{simon1977structure}, and \citep{amarel1983problems}).
Nijstad et al (2010) \citep{nijstad2010dual}, proposed a dual process theory which states that creativity can arise through two pathways: a flexibility pathway and a persistence pathway.

Many authors agree on the importance of both generative and evaluative processes and on the prominence of the interaction between them during creative thinking. As a consequence, the most promising approach is to rely on an integrated dual-process model of creativity which takes into account the different interaction mechanisms involving both S1 and S2 systems as well as both the generative and evaluative processes. 

An interesting theoretical model for the design of creativity-enhancing interfaces has been described in \citep{tubb2014}.  It is based on a two stage model of creativity described by Guilford, which characterises the creative process as a combination of ``convergent'' and ``divergent'' thinking, and the  dual process theory. The model is referred to as ``four quadrant model''.
It analyses the process by which a set of features parameters is changed to achieve a creative result (navigation of a finite parameter space). 

In the following we introduce the adopted rationale for the integration of a creative module inspired to the four quadrant model in the PSI cognitive architecture.  

\begin{figure}[htbp]
\begin{center}
\includegraphics[width=0.9\textwidth]{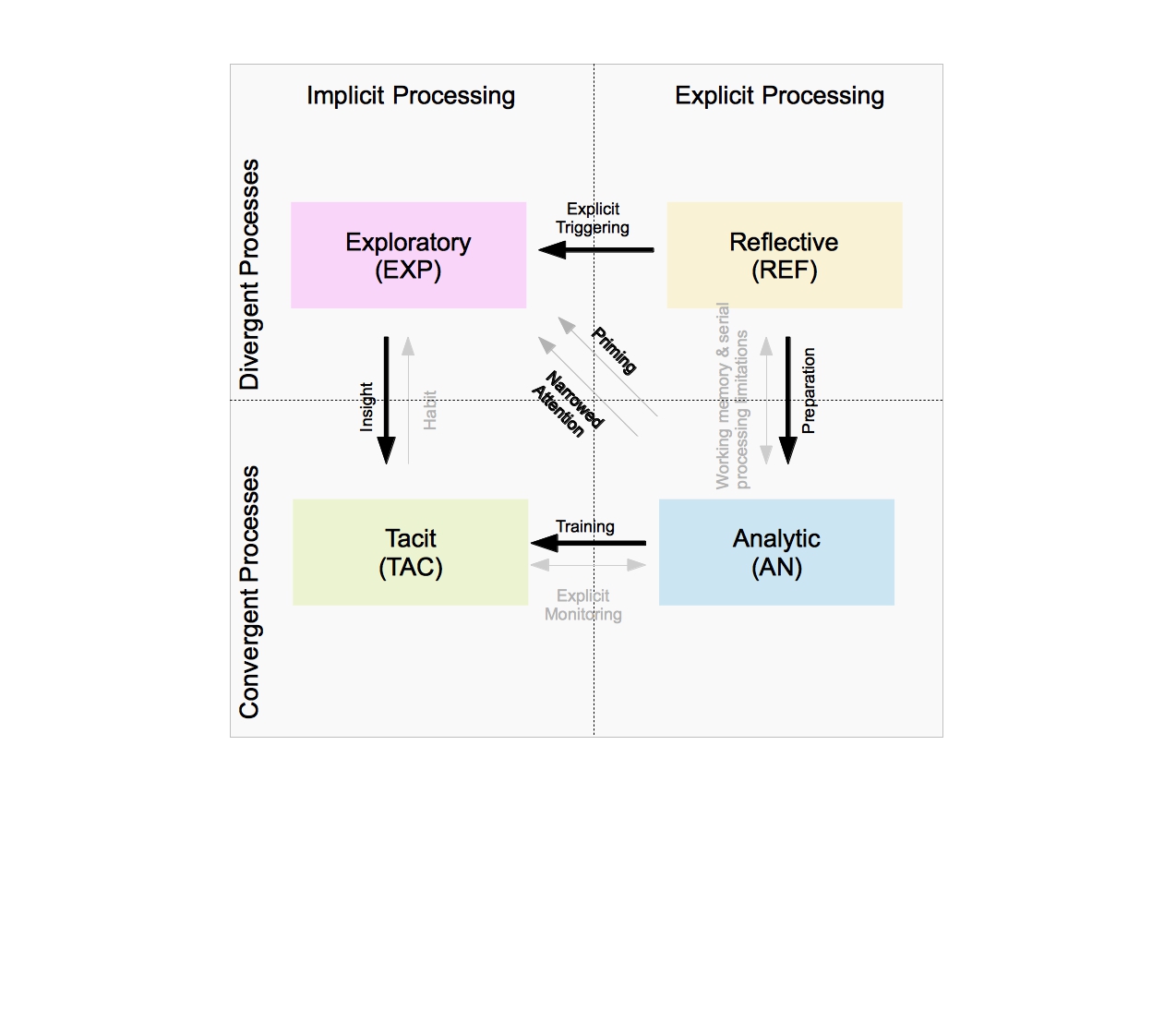}
\caption{Four quadrants model \citep{tubb2014}. We have highlighted only the interactions  involved in the creative process proposed in our system.}
\label{4quadranti}
\end{center}
\end{figure}

According to this model, both S1 (implicit processing), and S2 (explicit processing)  may: i) conduct convergent or divergent strategies ii) generate new ideas (new solution to a problem) and iii) converge to the more appropriate solution as illustrated in figure \ref{4quadranti}. 

S1 can generate new ideas  (EXP quadrant) by means of associative, combinatorial or transformational processes that can be therefore evaluated by S2 through an analytic strategy (AN quadrant), but  it can also  perform a Tacit strategy to converge to the more appropriate solution, using  knowledge formed in the past  to quickly access default ``previous best'' behaviour (TAC quadrant). Moreover S2 can perform a divergent, reflective strategy  to intentionally generate very distant points in solution space (REF quadrant). 

The interactions among the the four quadrants  can contribute to the generation of  creative solutions, but that can have inhibition effects, as discussed in  \citep{tubb2014}. In the figure we have highlighted  the interactions  involved in the creative process proposed in our system.

We believe that the application of the four quadrant model can be useful to extend the creative capabilities of existing artificial creative systems towards a more realistic and cognitively grounded scenario. 


\section{Cognitive architecture}
\label{sec:arch}
The proposed cognitive architecture is inspired to the PSI model \citep{bach2009principles}: we have made  small modifications to some functions and processes blocks to the original PSI model in order to implement the  S1-S2 model and its four quadrant interpretation (fig. \ref{4quadranti}). 
A creative process arises from a set of interactions highlighted by bold arrows in fig. \ref{4quadranti} involving  S1 and S2 systems operating in both convergent and divergent processes.
Such interactions can be described in terms of a dual theory model, iterating between divergent and convergent thinking, and embracing  all the four kinds of processes.
Each process is labelled according to the associated kind of system: for example a S1 system operating a divergent process is executing an ``Exploratory'' task.

The creative process starts if the agent has an adequate motivation. It is inspired by the perceptions of the agent (perception module), and it is carried out through phases of exploration, evaluation and, if necessary, reflection and re-planning. These stages require  access to the LTM and the working memory of the agent and different cognitive processes. 
The resolution level modulates the analysis of the perceptions, the access to the memory and the activation of the different cognitive processes. The creative process leads to the imagination of an artwork. However the mental representation of an artwork does not suffice   to the creative process. As a matter of fact the imagined artwork  must be  physically realized in order to explicate ideas in a tangible manner. 


In the next section we describe  the modules that are mainly involved in the creation of an  artwork, the evolution of the creative process and the influence of the evaluation on the artist, while the other parts of the architecture related to the physical  execution of the artwork will not be discussed since they are beyond the scope of this paper and for this reason they are shaded in the figure.


\begin{figure}[htbp]
\begin{center}
\includegraphics[width=0.9\textwidth]{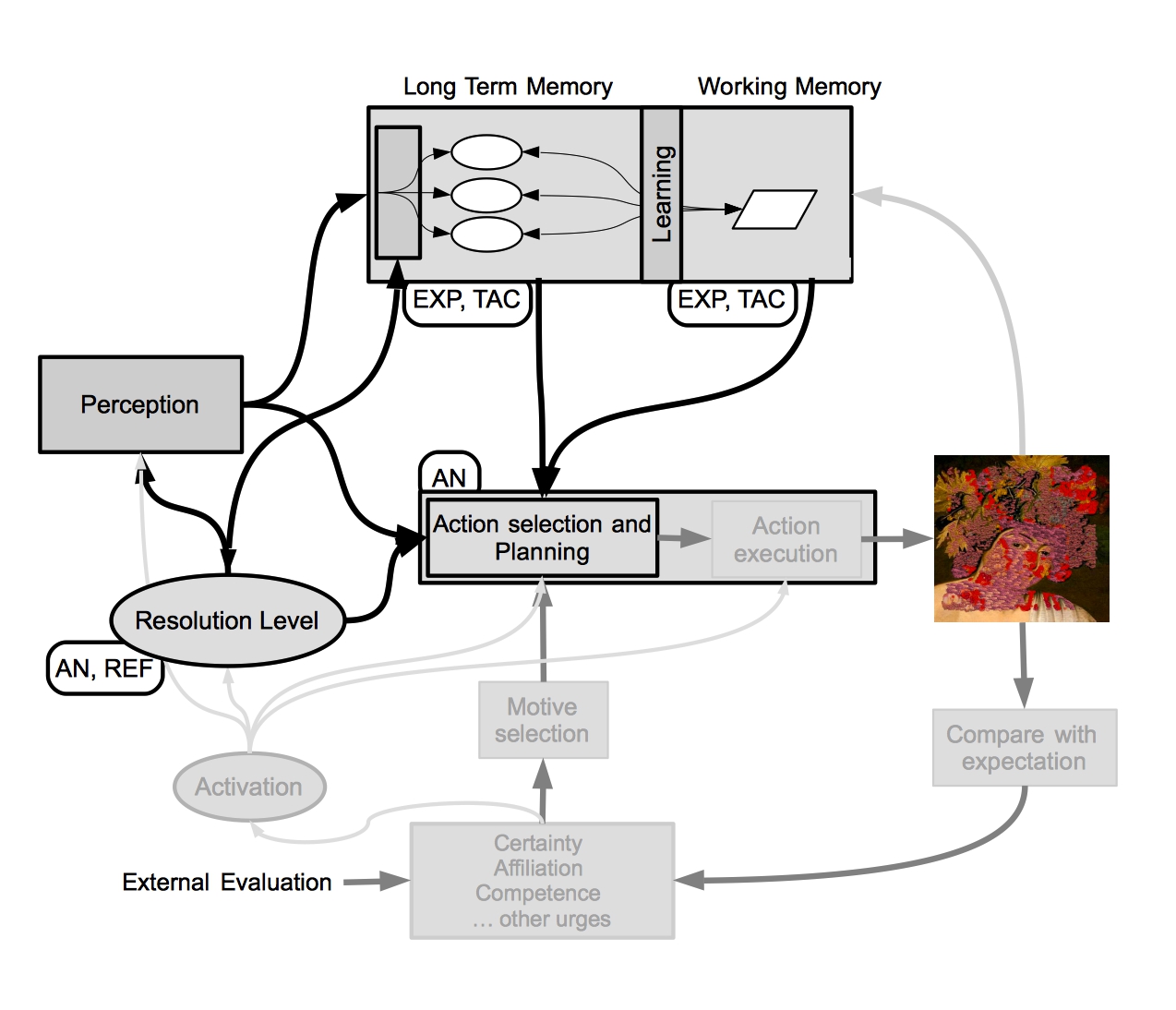}
\caption{The PSI architecture. The shaded blocks are not relevant for the current description of the dual process based creative path. The labels on each block refers to fig. \ref{4quadranti} and indicate the role of each block during the creative process.}
\label{schemaPSI}
\end{center}
\end{figure}

\subsection{Resolution Level}
\label{ssec:resl}
In the original description of the PSI model, the Resolution Level block operates both as a modulator working on the encoded percept (output of the sensing phase) and as communication module with the LTM.
In our perspective the resolution level can additionally control the S1 process outputs towards S2 processes. 

The resolution level is correlated in an inverse proportion with an activation parameter, which in turn is influenced by the urges of the artist \citep{bach2009principles}. 


The creative process starts when there is an adequate level of motivation, which leads to a value of Activation that motivates the artistic production. It is aimed to the achievement of a creative goal, like realizing an artwork. The resolution level takes part of the process when the system looks for an inspiration arising from its perceptions. A high level of Activation leads to a reduction of the resolution level. This could also lead to a speed up of the cognitive process since less details have to be taken into account. On the contrary, a lower level of Activation leads to a higher value of resolution, requiring a deeper analysis of the details \citep{bach2009principles}. 
In particular, the resolution level directly influences an exploratory parameter, which determines the distance from the cluster center of the working memory SOM (described in the next section) to consider different possible replacements of image regions. Furthermore, the resolution level controls also how much detailed should be the check of the result obtained by the S1 process from the S2 system.

The perceptive module output (the source of inspiration)  triggers the creativity process and the beginning of the exploratory process.
The resolution level (RL) controls the activation of both  S1 processes (tacit (TAC) and exploratory (EXP) in figure \ref{schemaPSI}). For the exploration process,  RL may decide to broaden or narrow the exploration regions, according to the urges that indirectly influence the RL. For what concerns  the activation of the tacit processes, RL may lead to the retrieval of default (e.g. previously learned) associations in the LTM. The unification of different types of S1 strategies follows the rationale introduced in \citep{lieto15ijcai} based on the assumption that different types of autonomous S1 processes need to be introduced and integrated in a cognitive architecture aiming at being a realistic model of human cognition. 
RL also controls the interaction between the Tacit Processes (TAC) and the Analytic ones (AN) during the  S2 checking of the results of the exploration process.
The Resolution Level is also involved  when a change of association domain is needed in order to try to satisfy the internal model of the agent w.r.t an internal model of the desired output.  This mechanism is considered as being reflective (REF).

%

\subsection{LTM and working memory}
\label{ssec:LTM}
\label{}
During the training phase the system learns and properly organizes its knowledge about  artistic notions and  the main execution techniques into the Long Term Memory (LTM). The term Long Term Memory (LTM) refers to a memory space that stores information for a long time-span (even a lifetime). This memory is recognized in all cognitive theoretical views \citep{Cowan2008}.
In our work the LTM is an associative memory that stores whole images or image details or features useful for a creative process  involving artwork or image creation.
This knowledge is aimed at representing image features clustered by subject. Information in this memory is organized in domains, were each domain stores information on objects of the same kind.

These image details are recalled by means of an association with the sensory input, or other memorized images and even with a conceptual representation (or internal representation) of the artwork involved in the creative process. The associative mechanism implies that the tasks involved can be tacit or exploratory as shown in fig.\ref{schemaPSI}.

In the presented architecture the LTM is constituted by a set of associative memory modules, each of them implementing a Domain: each Domain is a set of image features belonging to specific kind, like flowers, animals and so on.
Input from sensors are sent to the proper domain at the first level and they are memorised or completed when necessary as explained below. The second level contains the associations among different domains that will be further explained in the following and are related to the working memory.

The associative memory modules that we propose are inspired by the work reported in \citep{Morse2010} and are implemented using Self Organising Map (SOM) neural networks \citep{Kohonen2001}. Details on this architecture can be found \citep{augello2013}.

Self Organising Maps are neural networks constituted by a single layer of neural units usually organised in a 2D grid. After a successful training phase each neural unit ideally approximates the centroid of an input pattern cluster and the neighbour units represent similar values. This way each neural unit corresponds to a sort of average pattern for a cluster of inputs.

In the proposed architecture we also need a memory area for the artwork execution. The Working Memory (WM) can effectively identify this memory area used by the running processes. In this memory the pieces of information are stored for a short or medium time-span. In some cognitive theories this memory is not separated from the so-called Short Term Memory \citep{Cowan2008}. 
In our model the WM is used to temporarily store the associations among separated domains. These associations come from the similarity of two pieces of information in separated domains. 
The WM is built by using a large SOM that clusters all the image details and features in all the Domains. This will enable the substitution and exploration mechanism that is in the background of the creative process under consideration. 

The information coming from the domain modules is represented in the Working Memory by using more general features. For example, if a domain is used to memorise images of trees and one of the SOM in the  array of a Domain module memorises the shape of the leafs, the SOM in the working memory can use the dimension of the bounding box\footnote{the bounding box is the rectangle surrounding an image detail} as a feature. 
When we want to mix together objects from other domains we can consider objects that have the same bounding box.
The substitution will be driven by the second level SOM whose aim is to link the different domains by clustering the pieces of images memorised in the long term memory. 
A substitution according to the bounding box dimensions is a simple criterion but a more general set of features  could also be employed.

The WM SOM  implements the spreading of the  focus of attention since  it  mixes objects from different domains and group them just considering very rough characteristics. This module is used when the attention focus is relaxed and it is necessary to retrieve different objects and ideas. 
When completion is obtained by using ``parts" or memories that are outside of the domain of the original image, or input, we are making an association that is not casual.
This can happen when the recalled part is used to obtain memory contents from other, different domains. In this case, the associations are the ones memorised in the WM SOM, associations that involve two features of a different kind.

\begin{figure}[htp]
\centering
\includegraphics[scale=0.4]{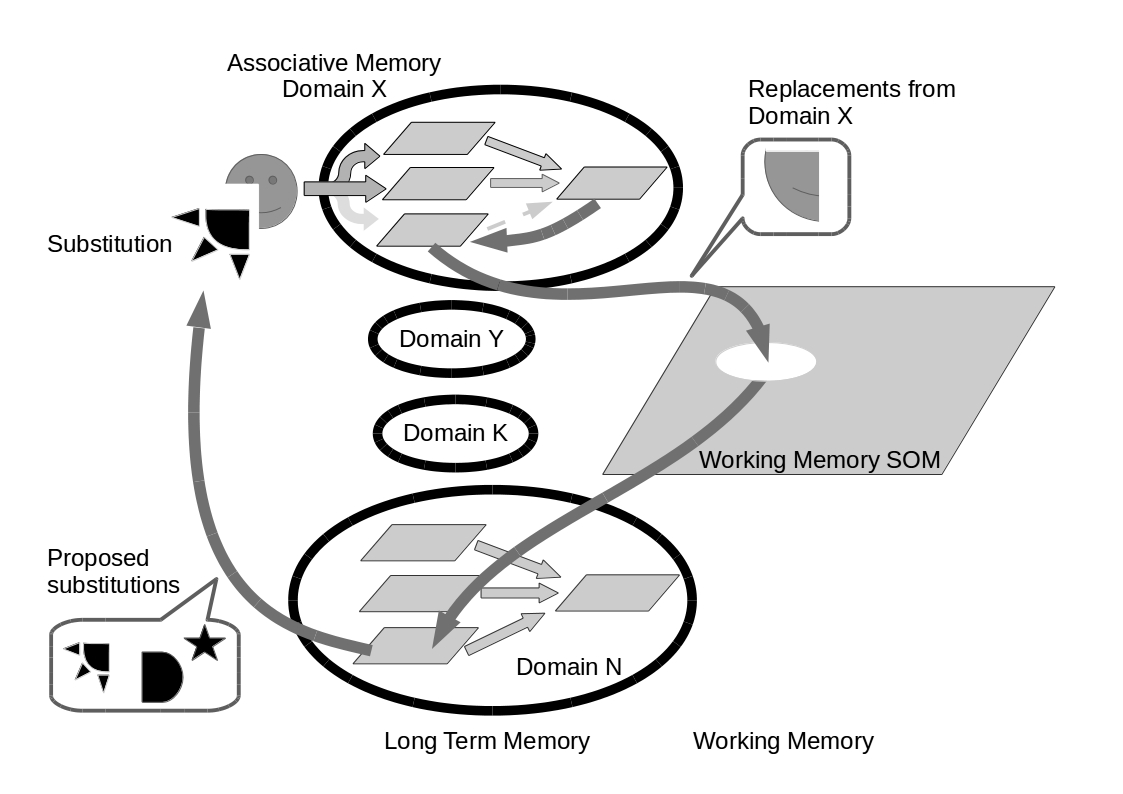}
\caption{The completion procedure outside the domain.}
\label{fig:CompletionOutsideDomain2}
\end{figure}

In fig. \ref{fig:CompletionOutsideDomain2} it is sketched the whole process: the missing part is recalled as said before however, in this case, it is not sent to the output but it is sent to the WM SOM where it is used for recalling objects from different domains.


The recovered information is used as a reference in order to obtain the missing part that is sent to the second level SOM. This signal excites a unit of the WM SOM and its output is sent back to all the associative memory of the other domains. Each domain answers with a list of the excited units that point out to a set of signal corresponding to the memorised objects. As indicated in fig \ref{fig:CompletionOutsideDomain2} all these objects are proposed as substitution of the missing part.
At this point the completion proposed from the original domain is again used as a reference: all the proposed substitution are compared to the original completion and the most similar one is chosen as a substitute.

\subsection{Evolution of the creative thinking}

In the following, we show some examples  of the system by explaining the cognitive flow.
%
 
First of all, we hypothesize that the artificial agent has some pre-defined tasks and domains. When motivations and urges determine the activation of the task \emph{make a portrait by association} $T_{pa}$, the sensing module,  the planning module, the resolution level, LTM, and STM are involved in a complex path of \emph{thinking} that switches from S1 to S2 processes and vice-versa.

The domain \emph{faces} $D_{faces}$ is pre-associated to task $T_{pa}$, but another domain needs to be chosen to have \emph{creative} association mechanism.  The agent takes inspiration by sensing its environment in order to choose a relevant domain: it looks for known objects, colors, and detect facial expressions of people (\citep{infantino2013feel}). This phase triggers the exploratory thinking (implicit divergent process EXP in figure \ref{4quadranti}) searching possible associations between two domains.  For example, in the case of the Arcimboldo painting style (\citep{iccc2014}), the flowers domain $D_{flowers}$ could be used to explore creative associations with $D_{faces}$. The exploration is also guided by tacit, implicit-convergent thinking TAC, and derives from a training process performed by analytic process (AN quadrant).

\begin{figure}[tb]
  \begin{centering}
  \includegraphics[width=15.00cm]{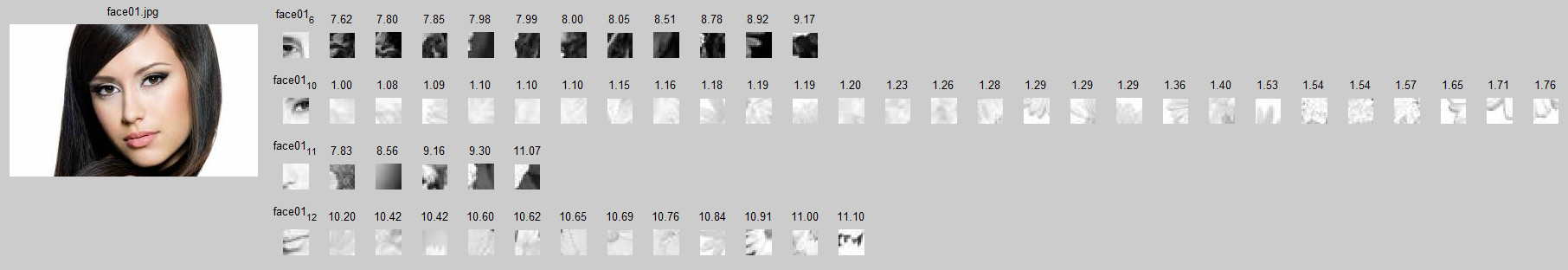}

  \caption{Example of possible candidates for replacement by association}
  \label{hubassociation}
    \end{centering}
\end{figure}

Figure \ref{hubassociation} shows an example of possible replacements by faces/flowers association mechanism. Four image regions extracted from original face image activate the hub-som clusters indicated in the figure. The element corresponding to the cluster centroids could be chosen for substitution, however we can set an exploratory parameter depending on distance from the cluster center to allow the system to consider other replacements. 

\begin{figure}[tb]
  \begin{centering}
    \includegraphics[width=0.45\textwidth]{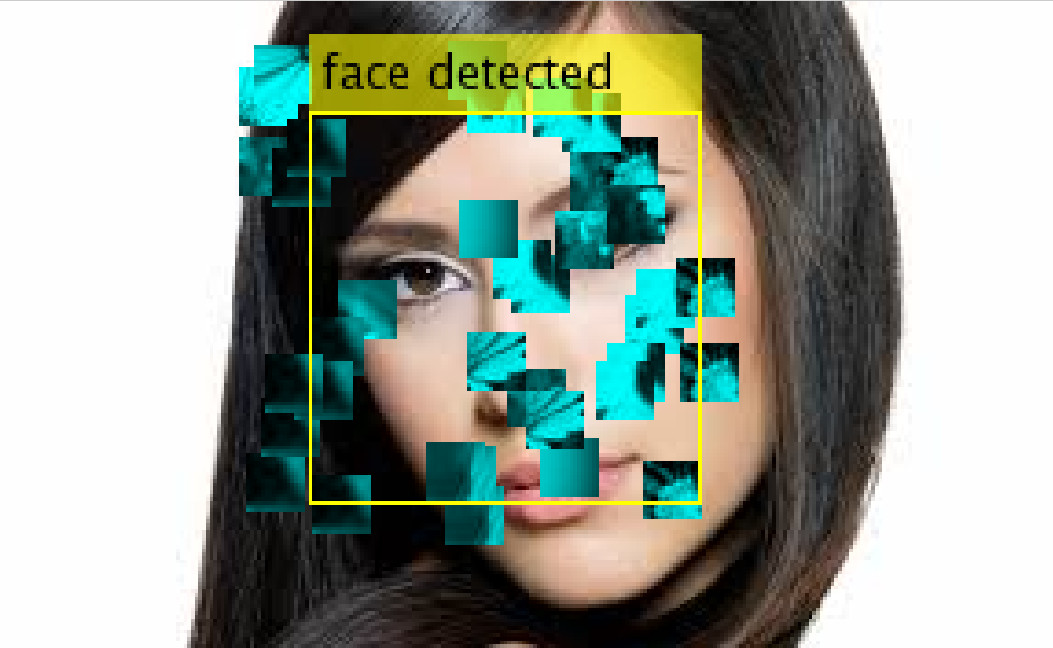}
    \includegraphics[width=0.45\textwidth]{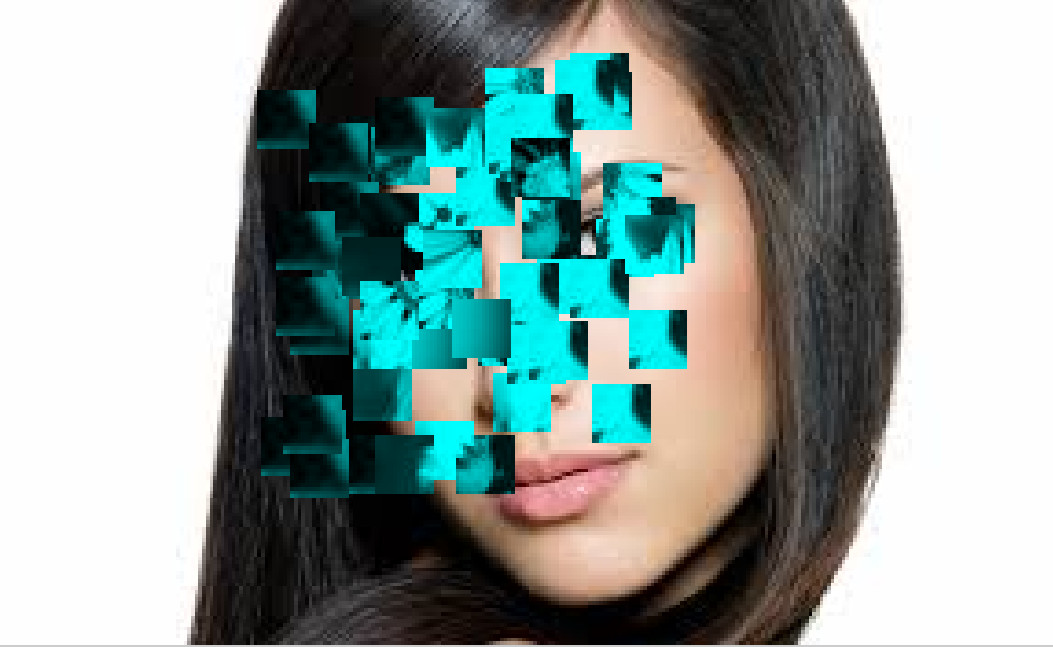}

  \caption{A face detection algorithm checks the presence of a recognizable face. The left side of the figure shows a positive check, while the right side shows a result  not valid for the artwork creation}
  \label{facedetection}
    \end{centering}
\end{figure}

Analytic component AN also monitors the mental execution of the artwork using the chosen association. The monitoring algorithm  is part of the preparation phase that arises from reflective thinking. In the case of faces-flowers association, for example we use  a face recognition algorithm that checks the presence of a recognizable face in the result of the substitution by association process in the mental image of the artwork (see example in \ref{facedetection}). Other relevant features  are RGB, CIE L*a*b*, HSV histograms, and two different texture representations (see section \ref{sec:expres}).

When the replacement is accomplished, the S2 system, performs a check of the  result obtained by the S1 process. The check is performed by means of an analytic strategy. 
In the specific case, the analytic strategy relies on a face detection algorithm which checks the presence of a face. 
Let us suppose that the result is the one shown in the left side of figure \ref{facedetection}, in this case the check returns a positive result and the system can execute its artefact. 
If the check is negative, as in the case shown on the right side of figure \ref{facedetection}, this leads to a reflective strategy. According to the reflective strategy the system can  trigger another explorative strategy, or can relax or modify the checking function used in the analytic strategy.

The reflective strategy is implemented by means of a set of rules in order to make the best decision according to the specific situation (the result obtained by the analytic strategy).

When the model of the painting is ready, the planning module starts the physical execution of the artwork. Analytic monitoring (AN) based on S2 processes, compares the product of execution to the internal model, and uses another internal evaluation function that realizes the internal expectation (a discussion about the creative process as an iteration between the conceptual representation and the final form could be found in \citep{Palle2015}). 

\begin{figure}[tb]
  \begin{centering}
    \includegraphics[width=4.5cm]{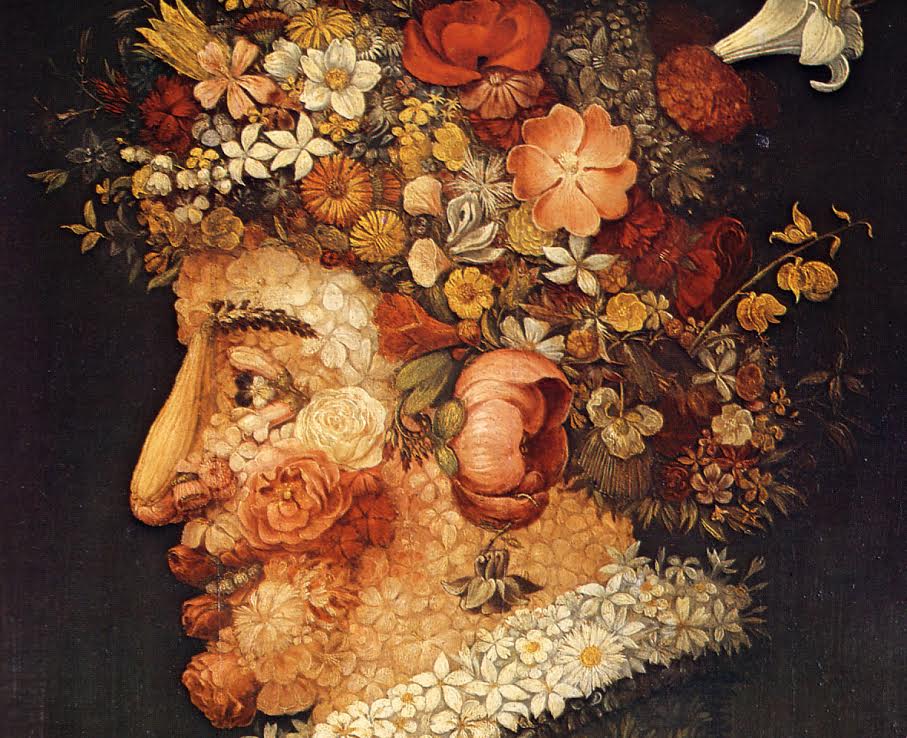}
    \includegraphics[width=4.5cm]{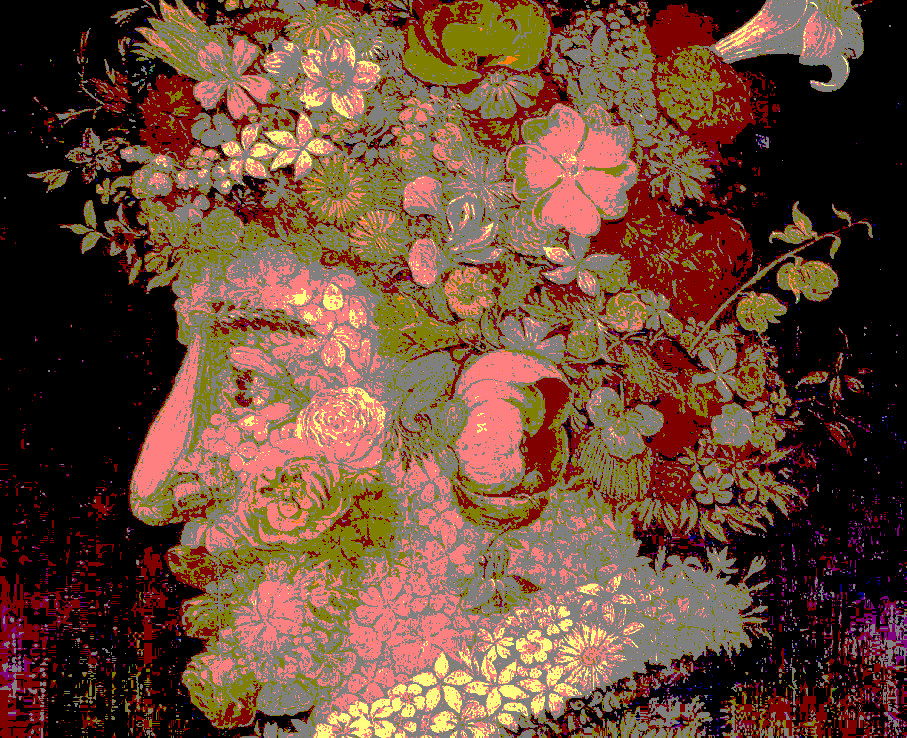}
    \includegraphics[width=4.5cm]{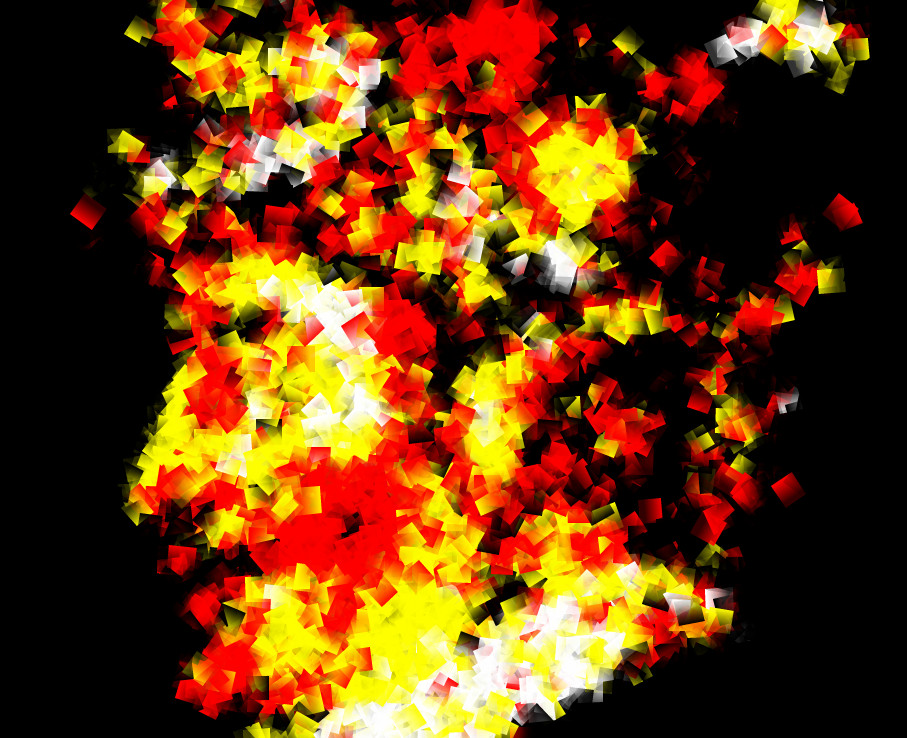}
 
  \caption{Different execution means can influence the creative results and even cause the replanning of the artwork creation}
  \label{expectation}
   \end{centering}
\end{figure}

For example, the internal model of the artwork could use a large set of colors (see example in fig.\ref{expectation} on the right), but in the execution phase the agent can use only few colors (see fig. \ref{expectation} on the middle) or a shape of paintbrush different from a dot (see fig. \ref{expectation} on the left). In the first case, the agent could accept the difference from its ideal model, but the second one could be rejected causing a replanning of artwork execution or creation. 

\begin{figure}[htbp]
\begin{center}
\includegraphics[width=0.9\textwidth]{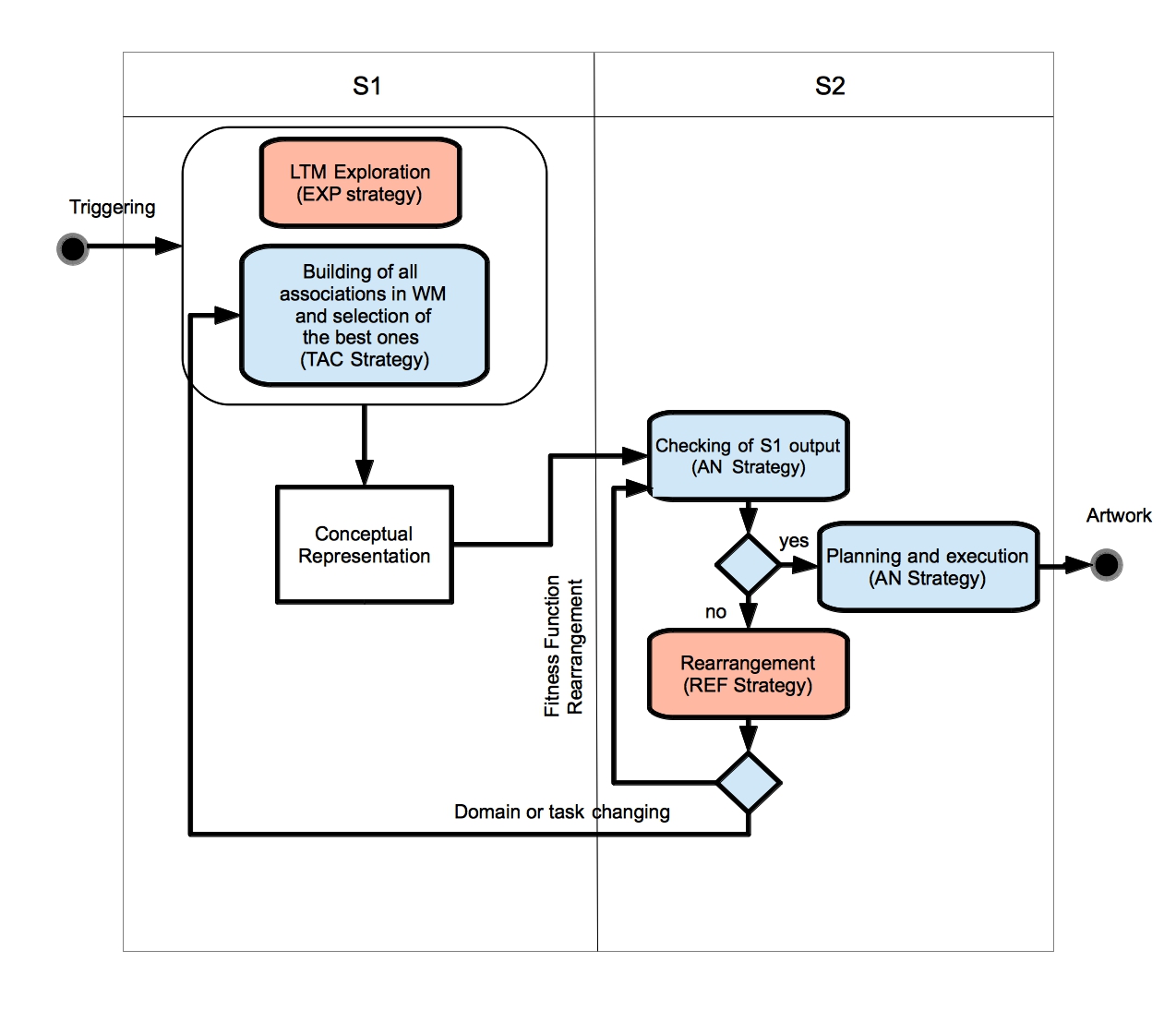}
\caption{Activity diagram of the creative thinking process}
\label{creativethinking}
\end{center}
\end{figure}

Moreover an external evaluation judges \citep{iccc2014} the results of the whole creative process and causes, when necessary, modification of the execution planning. Failures, i.e. unsatisfactory results measured by internal and external evaluation functions, may determine the modification of evaluation parameters in allowed ranges, or modification of execution mechanisms, or the change of domains for associations, or, in the final instance, the relinquishment to perform the task.
Figure \ref{creativethinking} describes the above explained creative  process as an activity diagram to better understand the switching between S1 and S2 processes that leads to artwork creation. In this diagram, the two evaluation phases are quite evident (indicated as diamond shapes), and they could activate a rearrangement of execution strategy by means of the REF process, or determine domain or task changing as previously stated.

\begin{figure}[tb]
\begin{centering}
	\includegraphics[width=4.5cm]{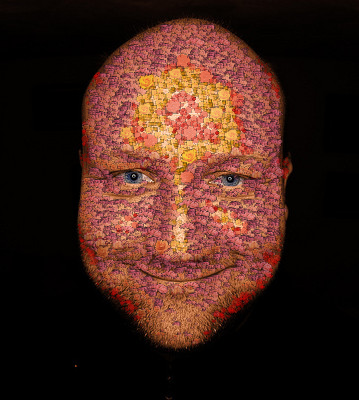}
\includegraphics[width=4.5cm]{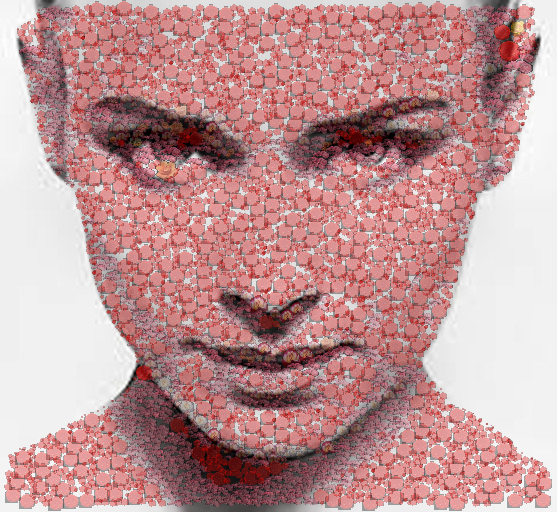}
\includegraphics[width=4.5cm]{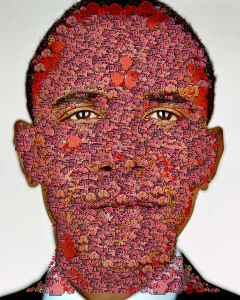}
\includegraphics[width=4.5cm]{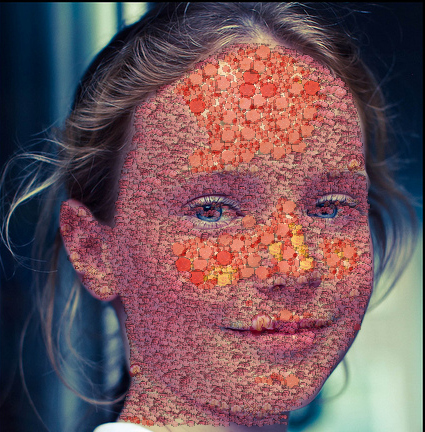}
\includegraphics[width=4.5cm]{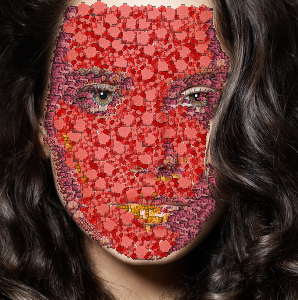}
\includegraphics[width=4.5cm]{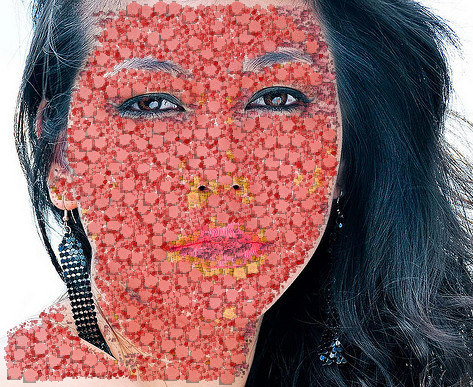}
	\includegraphics[width=4.5cm]{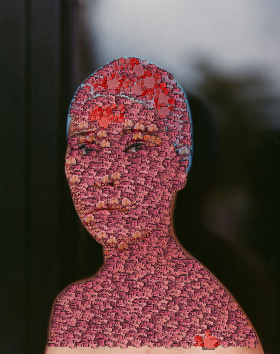}
\includegraphics[width=4.5cm]{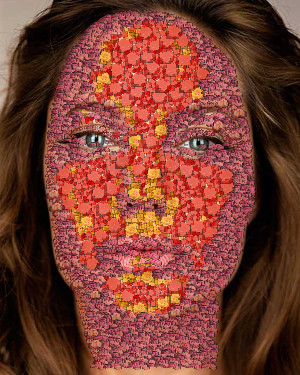}
 \caption{Some of the final obtained results}
\label{fig:others}
\end{centering}
\end{figure}

The advantage of framing the architecture in a dual process model is mainly that one of having a ``transparent'' description of the processes that occur, with the possibility of varying parameters such as the tendency to explore, the tendency to reflect, and so on. One possible objective to pursue in future work, might be to test different versions of agents focused (with varying degree) on some interactions (eg. agent A focused mainly on reflective thinking, Agent B on creative ``exploration'', and so on). Their behaviours and products could correspond to different creative styles. Basically, it could strengthen or inhibit the activation of some processes, and analysing if this makes a difference in the output. This opportunity of enhancing or inhibiting is possible thanks to the explicit description and representation of these processes in the model. In case of different results (which seems the most potentially interesting outcome) it would be interesting to do an analysis of the different types of output and trying to understand what are the processes that led to different behaviours.

\subsection{Motivation of the artist}

The motivation parameter is the element that activates the creative process and influences the execution planning. This parameter is changed through an evaluation process (both internal and external). In fact, a good evaluation of the  artworks and the receiving of  gratifications from the public increases the  motivation of the artist. At the same time, the motivation  increases when the artist, undertaking a difficult  activity,  obtains a valuable result (with respect to his expectations). The consequence is that it experiences a great satisfaction, feeling some sort of competence.

External and internal evaluation have a very important role determining the \emph{urges} of the artist and, consequently,  his motivation.  In particular, the evaluation influences  Competence and Certainty, and with the other urges it determines the motivation of the agent. \emph{Competence} is defined by Goertzel (\citep{cai2011openpsi}) as \emph{the effectiveness of the agent at fulfilling its needs}.  \emph{Certainty} is defined by Goertzel (\citep{cai2011openpsi})as \emph{the confidence of the agent's knowledge}.  In the specific case of a creative agent, we can define the Competence as the effectiveness at fulfilling correctly a set of creative acts, while the Certainty depends on the amount of styles judged as of \emph{pleasant} used by the agent. Here we focus on the cognitive urges, which influence what we call \emph{artistic motivation} or \emph{creative needs}.

During the execution of the artwork the artificial  artist compares what is  produced and the expectations with respect to its internal model. If the expectation of the agent is disregarded, and the outcome of an action does not reach the goal, something in the mechanism should be changed. Instead, a correct accomplishment of a technique allows the agent to achieve a positive evaluation, with the consequence that  the value of his competence increases, and consequently his motivation. Therefore in our model, the internal evaluation affects the value of  the Competence urge in MicroPsi.

The external evaluation regards a post-production evaluation of the  artworks and it is obtained asking a set of users to express their judgements.  Every time an user expresses a judgement about an artwork, his opinion has an impact on the execution techniques  used to produce that result. Therefore, the external evaluation  affects the Certainty of the agent and consequently influences the motivation of the artist.
The values of Competence and Certainty  determine a particular degree of the artist's evolution. As an example, a beginner is  an artist having both low Competence and Certainty, while an artist that has both high Certainty and Competence is an acclaimed artist. Different combinations of these values could be referred to other interesting states of the artist, and could lead the artist to reconsider its choices, by means of a reflective strategy.


%
%



\section{Experimental details and final results}
\label{sec:expres}

In this section we highlight some relevant details on the experimental process. 

The input of the process is the image perceived by the visual sensors and ``inspires'' the system triggering the creative task. Not all the parameters are known a priori; many of them are related to the current status of the system. As a consequence, the output generated by the system is different according the status of its own at the beginning of the process; furthermore, the creative process depends also on the ongoing output.
Here we point out the process leading from the perception of a face to the creation of an Arcimboldo like artwork.

The face domain is processed with an alternative domain that is chosen according the particular inclination of the system (considered the past experiences). A particular case is shown where the agent deals with the flower domain. The selection of the domains has been described in previous paragraphs.

The images and characteristics of the flowers are used to substitute and replace portions of the image. They are taken from an initial inspiration, or they are used to compose a whole artwork where the parts are given by previous constraints. They are chosen according a final desired output as shown in Figure \ref{fig:CompletionOutsideDomain2}.

 The manner in which the flower patches are used to replace pieces of faces is realised through the hierarchical process illustrated in section \ref{ssec:LTM}. The result is the substitution of part of the initial image with patches that are as similar as possible to the original regions to substitute.
 
 The similarity is evaluated by mapping portion of image on SOMs dealing with the specific domain taking into account the information regarding color and texture. The color information is evaluated according histograms based on a representation in color space while for texture, spatial frequency and pattern detecting features are considered.

For each image five different features are extracted. Three of them are referred to color information and two of them to texture information.
For color information, the representation of visual data in three different color space is taken into account. The color space are RGB, CIE L*a*b* and HSV.
The RGB features are derived from the histograms of the spectral components or \textit{red}, \textit{green} and \textit{blue}. The model is based on a Cartesian coordinate system and the color space can be represented as a cube where each point correspond to a given color.
To represent color content a vectorization is adopted. To each channels are associated ten bins that are used to store the color data for a given range of values. The same representation is used for all the three channels. The RGB representation is the typical representation of color images so the feature extraction is straightforward.
other chromatic space are also used to capture multiple aspects of the chromatic perception.

The RGB color space is connected with the perceptive capability of the human visual system that is mainly sensitive towards these three channels. The HSV space is used to have a representation that is more similar to the way humans describe colors and are suitable for natural language interpretation. When we describe an object we use the concepts of hue, saturation and brightness. Hue is a color property that describes a pure color (pure red, pure yellow and so on...). Saturation gives a measure of the degree to which the pure color is diluted with white color. Value is bound to a subjective descriptor that is difficult to describe. It embodies the achromatic notion of intensity and is one of the key factors in the description of the color sensation. This model has the further advantage to split the color information (hue, saturation) from the intensity information that is very similar to the gray scale representation of a given image. Also in this case ten bins are chosen for each channel. They are used to store how many pixels fall in a given range of values. Concatenating the histograms for the three channels is obtained a representation for the global color information \citep{Smith_HSV}.

The CIE L*a*b* color space is a representation conceived to include all the perceivable colors, which means that its gamut exceeds those of the RGB and CMYK color models . One of the most important properties of the L*a*b* model is that it is device independent. It means that the colors are defined independently from their generation and from the device they are displayed on. The value of L represents the lightness of the color that ranges from black to white. The a* value represents the color in the axis that represents the red to green line. The negative a* values are referred to green while the positive values of a* values are referred to red. The b* value represents the color along the yellow to blue variation. Negative b* values correspond to blue whileï¿œ positive b* values correspond to yellow\citep{Hunter_LAB}.

For the texture information we characterize the image according two features that are given by Gabor Filters and Haar Filters.
The Gabor Filters are filters that capture frequency and orientation of the visual patterns \citep{Gabor_Filters}.ï¿œ In the spatial domain, a 2D Gabor filter is a Gaussian kernel function modulated by a sinusoidal plane wave. The filters capture the presence of patterns with different orientation and scales. The orientations are given by the direction of the sinusoidal wave while the scale is bound to the values of the sigma parameter of the Gaussian function. The transformations produce a multi-channel representation driven by orientation and scale in line with the Daugman theory that Human Visual System elaborates set of image filtered with narrow orientation channels (\citep{Daugman_Cortex}). The energy associated to each filter is represented through a value so considering four values of the sigma parameter and three values for the direction is obtained a vector with twelve elements
The Haar filters are formed by a set of bidimensional square shaped functions that form a wavelet family of a basis. 
They can be seen as a special case of Debauchies wavelets \citep{Daubechies:1992:TLW:130655}. The response to each filter is given by the integral of the grayscale image in a square shared region subtracted by the integral of the image in the complement of the given region. The shape and the number of subcomponent of the region are bound to the complexity of the pattern that is detected. The key advantage of a Haar-like feature over most other features is its calculation speed. Due to the use of integral images, a Haar-like feature of any size can be calculated in constant time.
For the considered representation we have taken into account twelve regions from the simplest to more complex, to capture different shape of patterns and with a global dimensionality that is equal to Gabor filter descriptors.


In figure \ref{fig:others} (reported at the end of the paper) some example of the obtained results arising from the creative process are reported.


\section{Conclusions and future works}

We have presented a cognitive architecture aimed at computational creativity by exploiting a dual process approach.
In particular, the architecture, based on a microPsi cognitive model, has been embedded into an artificial digital painter.
The creative process involves both divergent and convergent processes in either implicit or explicit manner. This leads, according to the four quadrant model,  to four activities (exploratory, reflective, tacit, and analytic) that, triggered by urges and motivations, generate creative acts.
These creative acts exploit both the LTM and the WM in order to make novel substitutions to a perceived image by properly mixing parts of pictures coming from different domains, that are already stored in the memory of the agent.
In the paper we have highlighted the role of the interaction between S1 and S2, modulated by the resolution level, which focuses the attention of the agent by broadening or narrowing the exploration of novel solutions, or even drawing the solution from a set of already made associations. 

In the near future we plan to extend the evaluation of our system by increasing both the complexity of the task to be solved (i.e. by adding additional artistic domains that can be potentially chosen by the agent) and by comparing our BICA-based system with not biologically-inspired ones solving the same problem with different strategies (by using, for example, random selection choices or rule-based mechanisms). 

Our expectation is that, in the final evaluation performed by human ``judges'', the ``creativity'' choices of our system, given its cognitive grounding, should be rated higher w.r.t. that ones provided by artificial systems adopting standard decision procedures.

As above mentioned, one of the next steps of the current research will be that one of investigating the creative output of the artificial agent by adopting different creative styles in different scenarios (e.g. the ``explorator'' agent can be compared with the ``analytical'' one and so on).

In this perspective, we also plan to investigate the adaptability of the present model w.r.t. the novel Stanovitch's tripartite framework \citep{stanovich2009distinguishing}. Such a framework belongs to the ``dual-process theories'' family of cognitive models and differs from other dual-process theories in its description of  System 2.  In particular:  processes usually ascribed to System 2 are assumed to belong to two types of systems, respectively called the ``Algorithmic Mind'', responsible for cognitive control, and the ``Reflective Mind'', responsible for deliberative processes.

According to Stanovitch, the division of human cognition into three sets of  processes, instead of the traditional two of dual-process theories, provides a better account of individual cognitive differences (e.g. individual difference with regard Algorithmic Mind processes are linked to cognitive abilities and fluid intelligence, while Reflective Mind differences are observed in critical thinking skills). Since we aim at precisely investigating  in which way the output produced by different types of creative agents depends from their individual creative style,  the analysis of the compatibility between the proposed dual process based model of computational creativity (in its 4 quadrant interpretation) and this new theoretical framework seems to be promising and, in our opinion, deserves further investigation.

\section{References}



\bibliographystyle{elsarticle-harv} 
\bibliography{csr_new}


%
%
%
\end{document}